\documentclass[10pt,twocolumn,letterpaper]{article}
\pdftrailerid{}
\usepackage{arxiv}
\usepackage[T1]{fontenc}
\usepackage{fix-cm}
\usepackage{microtype}
\usepackage{graphicx}
\usepackage{booktabs}
\usepackage{enumitem}
\usepackage{multirow}
\usepackage{amsmath}
\usepackage{amssymb}
\usepackage{makecell}
\usepackage{xcolor}
\usepackage[numbers,sort&compress,square]{natbib}
\usepackage{caption}
\usepackage{placeins}
\usepackage{flushend}
\usepackage[hidelinks,hyperfootnotes=false]{hyperref}
\usepackage{xurl}
\renewcommand{\headeright}{}
\renewcommand{\undertitle}{}
\renewcommand{\authorboxwidth}{0.99\linewidth}
\title{\textbf{MXCode: Vector Quantization for MX-Native Inference at Scale}}
\author{{
\renewcommand{\arraystretch}{1.3}
\begin{tabular}[t]{@{}c@{}}
\textbf{Gaurav Kolhe}\textsuperscript{1,}\thanks{Equal contribution.}\quad
\textbf{Cansu Demirkiran}\textsuperscript{1,*}\quad
\textbf{Shubhi Tyagi}\textsuperscript{1}\quad
\textbf{Sulaiman Sadiq}\textsuperscript{1}
\\
\textbf{Boyao Wang}\textsuperscript{2}\quad
\textbf{Richard Wang}\textsuperscript{1}\quad
\textbf{Maximiliana Behnke}\textsuperscript{1}\quad
\textbf{Shreyas Labhsetwar}\textsuperscript{1}
\\
\textbf{Levi Melnick}\textsuperscript{1}\quad
\textbf{Matthew Mattina}\textsuperscript{1}\quad
\textbf{Partha Maji}\textsuperscript{1,}\thanks{Correspondence: \href{mailto:parthamaji@microsoft.com}{parthamaji@microsoft.com}.}
\\
\\
\textsuperscript{1}Microsoft\\
\textsuperscript{2}Carnegie Mellon University
\end{tabular}
}}
\date{}
\hypersetup{pdfauthor={},pdftitle={},pdfsubject={},pdfkeywords={},pdfcreator={},pdfproducer={}}

\begin{document}

\maketitle[{
\begin{abstract}
\fontsize{11}{12}\selectfont
Memory capacity and bandwidth are major bottlenecks in large
language model (LLM) inference, motivating weight compression
beyond conventional 4-bit quantization. Meanwhile, FP4 computation
is becoming a standard capability in modern AI accelerators,
with microXcaling (MX) formats enabling efficient W4A4 (Weights \& Activations in 4 bit) matrix
multiplication. Vector quantization (VQ) enables approximately 2-bit weight storage, roughly halving the storage cost of MXFP4 formats.
However, existing VQ approaches often rely on codebook representations that are not directly compatible with native FP4 computation.
We present \textbf{MXCode (MX-Native Codebook Quantization)}, a framework
that bridges ultra-low-bit VQ compression and MX-native FP4
computation. MXCode introduces three complementary strategies: PostScale, which quantizes
learned codebooks to MXFP4; PreScale, which extracts MX scales
before codebook learning; and PreScale+, which additionally
compresses these scales using VQ. Together, they explore
tradeoffs among quantization quality, scale overhead, and
codebook granularity.
All three strategies reconstruct MX-native W4A4  operands  without materializing intermediate high-precision weight tensors.
We evaluate MXCode on three industrial-scale LLMs, including
Qwen3 (235B), DeepSeek-R1 (671B), and GLM-5.2 (753B), using the
Microsoft MX Emulation Library\footnotemark{} to assess
model quality under MX-compatible numerics. MXCode achieves
approximately 2 bits per weight
while maintaining competitive model quality, with workload-dependent degradation.
In GB300 serving experiments with Qwen3 (235B) model,
MXCode approximately halves model-load memory relative to Native FP4
while sustaining near-Native-FP4 serving throughput, and it reduces P90
TTFT by up to 94.5\% under memory-constrained prefix reuse compared to Native FP4.
These results, demonstrated through software-based reconstruction on existing FP4 hardware, inform the ongoing development and evaluation of MXCode for future MX-native inference systems.
\end{abstract}
\bigskip
}]
\footnotetext{\url{https://github.com/microsoft/microxcaling}}

\section{Introduction}
Large language model (LLM) inference is increasingly constrained by
memory capacity and bandwidth.Model weights drive accelerator memory requirements, replication density, and capacity for runtime state such as the KV cache. During autoregressive decoding, repeated weight transfers can dominate memory traffic, particularly at low-to-moderate batch sizes. While 4-bit quantization reduces these costs,
further compression is desirable for efficient large-scale serving.

Sub-4-bit compression must contend with hardware support centered
on FP4. Modern AI accelerators, including Microsoft
MAIA~\citep{maia26}, NVIDIA
Blackwell~\citep{blackwell}, and AMD Instinct
MI350-series~\citep{mi350}, support native MXFP4
(microXcaling) compute~\citep{mxspec23}. These formats combine low-precision elements with shared block scales to enable efficient FP4 computation. However, storage
precision need not match compute precision. Storing weights below four
bits per element (BPE) while retaining native FP4 execution can reduce
model footprint and HBM traffic without sacrificing low-precision
compute.

Vector quantization (VQ) enables ultra-low-bit weight compression by representing weight vectors with compact codebook indices, achieving approximately 2 BPE or lower. However, conventional VQ approaches typically employ FP16 or INT8 codebooks without explicitly considering MX block-level scaling~\citep{aqlm24,quip24,gptvq25,vptq24}. Some approaches also use large codebooks, such as 65,536 entries in AQLM~\citep{aqlm24}, increasing lookup footprint and potentially limiting on-chip caching. Mapping these representations onto MXFP4 hardware requires either quantizing codebooks into MX-compatible formats, potentially introducing block-scale distortion, or reconstructing higher-precision weights before conversion, increasing data movement and execution overhead. Ultra-low-bit compression for MX-native inference is therefore a hardware--software co-design problem involving quantization quality, codebook footprint, scale handling, and operand reconstruction. Furthermore, some VQ methods require gradient-based fine-tuning through codebooks, complicating deployment of large-scale pretrained models with frozen weights~\citep{aqlm24,quip24}.

\begin{table*}[t]
    \centering
    \caption{Block-scaled FP4 formats compared in this work. All use E2M1 elements and 8-bit block scales. The reported ranges refer to block scales, and BPE excludes tensor-level scaling overhead.}
    \label{tab:fp4}
    \begin{tabular*}{\textwidth}{@{\extracolsep{\fill}}lcccccc@{}}
        \toprule
        Format &
        Block size &
        Scale &
        \makecell{Min. positive\\scale} &
        \makecell{Max. finite\\scale} &
        \makecell{Optional tensor\\scale} &
        BPE \\
        \midrule
        E8M0-B32-FP4 &
        32 & E8M0 &
        $2^{-127}$ &
        $1.70\times10^{38}$ &
        -- & 4.25 \\
        E4M3-B16-FP4 &
        16 & E4M3 &
        $2^{-9}$ &
        $448$ &
        FP32 & 4.50 \\
        E5M3-B16-FP4 &
        16 & UE5M3 &
        $2^{-17}$ &
        $114688$ &
        FP32 & 4.50 \\
        \bottomrule
    \end{tabular*}
\end{table*}

MXCode addresses these challenges through an MX-native VQ framework combining approximately 2-bit weight storage with native W4A4 computation. Its three strategies, PostScale, PreScale, and PreScale+, jointly explore codebook quantization, MX scale extraction, and scale compression to balance numerical fidelity, storage efficiency, and execution compatibility. MXCode uses one-shot, gradient-free post-training quantization on frozen pretrained weights, avoiding retraining or backpropagation through codebooks and easing adoption for large-scale and third-party LLMs.

Our main contributions are:
\begin{itemize}[leftmargin=0.5\leftmargini]
    \item \textbf{MX-native vector quantization.} We introduce MXCode,
    combining approximately 2-bit weight storage with native W4A4
    computation through FP4-compatible codebooks and operand
    reconstruction without intermediate high-precision weights.

    \item \textbf{Compression strategies and large-scale validation.}
    We develop PostScale, PreScale, and PreScale+, exploring
    storage--quality trade-offs across three industrial-scale LLMs
    (235B--753B parameters).

    \item \textbf{GPU serving demonstration.} On NVIDIA GB300 GPUs,
    MXCode approximately halves model-load memory while sustaining
    near-Native-FP4 throughput, enabling $2\times$ data-parallel
    replication at comparable memory headroom and reducing P90 TTFT
    by up to 94.5\% under prefix reuse compared to Native FP4.

\end{itemize}

\section{Background and motivation}

\subsection{MicroXcaling (MX) FP4 Formats}

MicroXcaling (MX) enables efficient low-precision computation through fine-grained shared scaling and
low-precision element formats~\citep{mxformats23,mxspec23}.
MX formats continue to evolve, with the release of OCP MX
v1.1.0 in October 2026 reflecting ongoing industry-wide
standardization of low-precision computation~\citep{mxspec26}.

For a block of $B$ elements, each value is approximated as
\begin{equation}
    \hat{x}_{b,i} = s_b q_{b,i},
    \qquad i=1,\ldots,B,
\end{equation}
where $q_{b,i}$ is a low-precision element and $s_b$ is the shared
block scale. With element width $b_e$ and block-scale width $b_s$,
the effective storage cost, excluding tensor-level metadata, is
$\mathrm{BPE}=b_e+b_s/B$. Smaller blocks enable finer adaptation
to local value distributions but increase scale overhead.

We consider three block-scaled FP4 configurations,
collectively referred to as MXFP4 formats throughout
this paper: E8M0-B32-FP4, E4M3-B16-FP4, and
E5M3-B16-FP4. Among these, E8M0-B32-FP4 is
standardized in the current OCP MX v1.1.0 specification,
whereas E4M3-B16-FP4 and E5M3-B16-FP4 represent
emerging extensions to the MXFP4 design space.
We discuss their details in Table~\ref{tab:fp4}.

\paragraph{E8M0-B32-FP4.}
The OCP MX specification defines MXFP4 using 32 E2M1 elements
with a shared 8-bit E8M0 scale~\citep{mxspec26}.
Its power-of-two scaling provides a wide dynamic range and low
scale overhead (4.25 BPE), but no fractional scale precision.

\paragraph{E4M3-B16-FP4.}
NVIDIA's NVFP4 format groups 16 E2M1 elements under an FP8 E4M3 block scale~\citep{nvfp425}, enabling finer local adaptation through smaller blocks and fractional scale precision at 4.50 BPE. A tensor-level FP32 scale $g$ extends the effective range,
so that $\hat{x}_{b,i}=g s_b q_{b,i}$.

\paragraph{E5M3-B16-FP4.}
Unsigned E5M3 (UE5M3) reallocates the sign bit of the block scale
to exponent range while retaining three fractional bits
~\citep{finer26,ue5m326}. Its wider range mitigates
scale underflow with 16-element blocks without requiring an
additional tensor-level FP32 scale. The effective cost remains
4.50 BPE.

\subsection{Vector Quantization (VQ)}

While FP4 microscaling enables efficient low-precision
computation, its storage cost remains above four BPE. VQ enables
sub-4-bit compression by representing groups of weights
through compact codebook indices rather than independently
quantized scalar values~\citep{aqlm24,quip24}.

Given a vector $\mathbf{x}\in\mathbb{R}^{d}$ and a
codebook $\mathcal{C}=\{\mathbf{c}_1,\ldots,\mathbf{c}_K\}$
containing $K$ vectors of dimension $d$, VQ selects
the nearest codeword:
\begin{equation}
    j^* = \arg\min_{j\in\{1,\ldots,K\}}
    \|\mathbf{x}-\mathbf{c}_j\|_2^2,
    \qquad
    \hat{\mathbf{x}}=\mathbf{c}_{j^*}.
\end{equation}
Each vector is stored using a codebook index requiring
$\lceil\log_2 K\rceil$ bits, giving an effective index
cost of
\begin{equation}
    \mathrm{BPE}_{\mathrm{index}}
    = \frac{\lceil\log_2 K\rceil}{d},
\end{equation}
excluding codebook and scale storage.
By learning representative patterns
across multiple weights, VQ exploits inter- element structure that scalar quantization cannot capture.

Conventional VQ minimizes weight-space reconstruction error, although
errors in different weight directions do not have equal effects on the
layer output. Prior work uses Hessian-aware VQ~\citep{gptvq25}, which estimates weight
importance from representative calibration activations. Given a weight
matrix $\mathbf{W}$, its approximation $\widehat{\mathbf{W}}$, and
calibration inputs $\mathbf{X}$, codebook learning minimizes
\begin{equation}
    \mathcal{L}_{H}
    =
    \left\|
        (\mathbf{W}-\widehat{\mathbf{W}})\mathbf{X}
    \right\|_F^2.
\end{equation}
The input second moment
$\mathbf{H}=\mathbf{X}\mathbf{X}^{\top}$ induces a weighted distance
that gives greater importance to directions that are more strongly
activated by the calibration data. Hessian-aware VQ therefore biases
codebook learning toward preserving weight directions with a larger
estimated effect on the model output.

\subsection{Bridging VQ and MXFP4}

Despite its storage benefits, conventional VQ lacks a direct, efficient execution path on native MXFP4 hardware. The challenge is converting compact codebook-based weights into MXFP4-compatible operands while preserving compression efficiency without introducing substantial reconstruction overhead.

Existing VQ approaches often rely on high-precision learned codebooks or reconstruction procedures that are not directly compatible with MXFP4 arithmetic. Integrating VQ with native FP4 computation therefore requires converting decoded codebook values into MX-compatible operands, potentially introducing additional data movement, scaling, and conversion overhead. Furthermore, conventional VQ optimization does not explicitly account for the numerical constraints imposed by MX block scaling, which can introduce additional quantization error when codebooks are converted to FP4.

These challenges motivate jointly considering codebook representation, scale extraction, and compression efficiency. The following section describes the MXCode compression pipeline designed around these requirements.

\begin{figure*}[t]
    \centering
    \includegraphics[width=\textwidth]{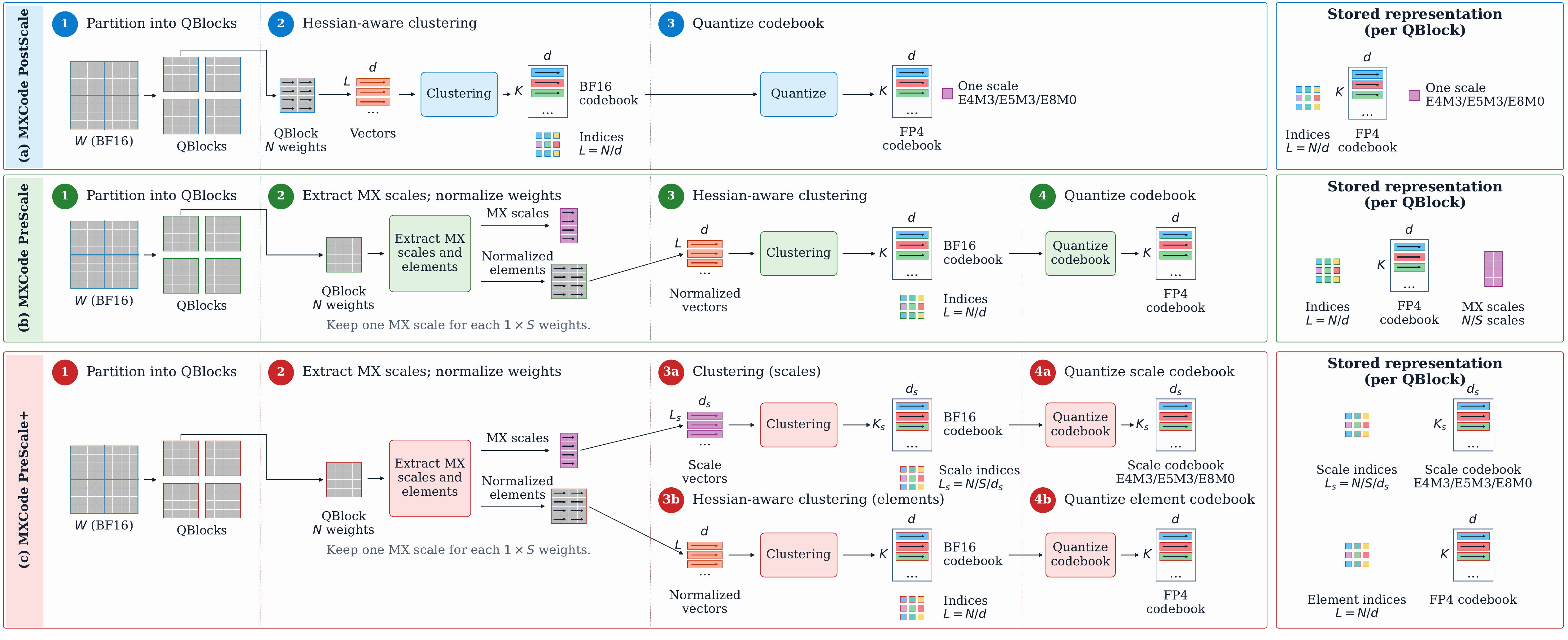}
    \caption{
    MXCode: MX-native codebook quantization pipeline; (a) PostScale, (b) PreScale, and (c) PreScale+ compression schemes.
                                        }
    \label{fig:mxpipe}
\end{figure*}

\section{MXCode Compression pipeline}
\label{sec:pipeline}

In this section, we introduce MXCode and its three schemes for integrating vector-quantized weights with MX-compatible FP4 computation: PostScale, which quantizes the learned codebooks using one scale per codebook; PreScale, which extracts fine-grained MX scales before codebook learning; and PreScale+, which reduces the scale overhead of PreScale through scale compression. These three methods are illustrated in Figure~\ref{fig:mxpipe}.

For all three schemes, starting from a BF16 weight matrix 
\begin{equation*}
\mathbf{W}\in\mathbb{R}^{C_{\mathrm{out}}\times C_{\mathrm{in}}},
\end{equation*}
we partition it into quantization blocks (QBlocks) of size
$B_{\mathrm{out}}\times B_{\mathrm{in}}$. Each QBlock learns an independent
element codebook with $K$ vectors of dimension $d$.
However, the three schemes differ in when MX scaling is applied relative to codebook learning, how the scales are associated with the FP4 codebooks, and how the resulting scale-storage overhead is represented, as described in the following subsections.

\subsection{\mbox{MXCode-PostScale}}
\label{sec:post}

Figure~\ref{fig:mxpipe}(a) illustrates \mbox{MXCode-PostScale}.
In this scheme, we
apply Hessian-aware VQ directly to the BF16 weights in each QBlock. Resulting codebook
$\mathcal{C}_V\in\mathbb{R}^{K\times d}$ is then quantized to an MX
representation by extracting one shared scale $s_{\mathcal{C}}$ and
quantizing all $Kd$ codebook values to FP4:
\begin{equation}
    \overline{\mathcal{C}}_V
    =
    s_{\mathcal{C}}\,
    Q_{\mathrm{FP4}}
    \left(
        \frac{\mathcal{C}_V}{s_{\mathcal{C}}}
    \right).
\end{equation}

Each QBlock
therefore stores an element-index stream, one FP4 codebook, and one
associated scale. During execution, the codebook scale is replicated
at the MX block granularity.

\mbox{MXCode-PostScale} introduces negligible scale overhead, at the cost of a much coarser scaling
granularity than that of standard MX formats: one scale is shared by all $Kd$ values in a codebook.
Smaller QBlocks improve codebook and scale locality, but increase
the amortized codebook cost and the number of codebooks managed by the
hardware.

\subsection{\mbox{MXCode-PreScale}}

Figure~\ref{fig:mxpipe}(b) illustrates \mbox{MXCode-PreScale}. Instead of
quantizing a codebook after it has been learned, we extract MX-style scales
from the weights before applying VQ.
We extract one scale per scale
block containing $S$ weights. For the scale-block size, we follow the native MX block size, $S=S_{\mathrm{MX}}$, e.g., $S_{\mathrm{MX}}=32$ for E8M0-B32-FP4.
For a scale block $\mathbf{W}_b$, we compute the scale for that block $s_b$ and block's normalized elements $\mathbf{V}_b$ as
\begin{equation}
    \begin{aligned}
    s_b
        &=
    Q_S\left(
        \frac{\lVert\mathbf{W}_b\rVert_{\infty}}
             {q_{\max}}
    \right), \\
    \mathbf{V}_b
    &=
    \operatorname{clip}
    \left(
        \frac{\mathbf{W}_b}{s_b},
        -q_{\max},
        q_{\max}
    \right),
    \end{aligned}
\end{equation}
where $q_{\max}$ is the largest magnitude of the FP4 element format
and $Q_S$ quantizes the high-precision scale to the target MX scale datatype, such as
E4M3, E5M3, or E8M0.
Hessian-aware VQ is then applied to the normalized
weights $\mathbf{V}$, and the resulting codebooks are stored directly
in FP4.

Unlike PostScale, PreScale retains fine-grained MX scaling, which comes with a higher
scale overhead that is independent of the QBlock size.
To reduce this scale overhead, $S$ can also be an integer multiple
of the native MX block size, $S=mS_{\mathrm{MX}}$. One stored scale is then shared across $m$ MX blocks and replicated per block during execution, with effectiveness evaluated in Appendix~\ref{app:scales}.

\subsection{\mbox{MXCode-PreScale+}}

Figure~\ref{fig:mxpipe}(c) illustrates \mbox{MXCode-PreScale+}, which
retains the fine-granularity scaling of PreScale while reducing its scale
overhead.
We extract MX scales before applying VQ as in PreScale and
then compress the resulting scale tensor using a second VQ
representation.
Neighboring scales are grouped into vectors
$\mathbf{s}_i\in\mathbb{R}^{d_s}$ and encoded with a scale codebook
$\mathcal{C}_S=\{\mathbf{c}^{(S)}_k\}_{k=1}^{K_s}$:
\begin{equation}
    a_i^{(S)}
    =
    \arg\min_k
    \left\|
        \mathbf{s}_i-\mathbf{c}^{(S)}_k
    \right\|_2^2.
\end{equation}
The scale centroids are then quantized to the target MX scale datatype, e.g., E8M0, E4M3, etc.
The normalized weights are separately compressed using Hessian-aware VQ and stored in FP4 similar to the PreScale scheme.

For all three MXCode schemes, codebook lookups produce FP4 elements and their
associated MX scales, which can be supplied directly to an
MX-compatible matrix-multiplication kernel during inference, without any reconstructed BF16
weight tensor being materialized.

\section{Results}
\subsection{Ablation Studies}
\label{sec:ablation}
In this section, we explore MXCode design space to identify effective configurations and guide design choices. Our ablations use Qwen3 (235B), sampling ten transformer blocks across model depth (layers 1, 10, 20, 30, 40, 50, 60, 70, 80, and 93). We evaluate the gate, up, and down projections of 10 randomly selected experts per block, totaling 300 weight matrices, and reuse this set across configurations for consistent comparisons. All experiments construct vectors along the input-channel ($C_{\mathrm{in}}$) dimension for hardware-friendly implementation and use signal-to-quantization-noise ratio (SQNR) as a proxy for model quality.

\subsubsection{VQ Parameter Selection}

We first evaluate the VQ dimension $d$ and codebook size $K$ without any MX scaling, using one codebook per weight tensor. The effective storage cost in terms of bits per element (BPE) is
\begin{equation}
    \mathrm{BPE}_{\mathrm{VQ}}
    =
    \underbrace{
        \frac{\lceil \log_2 K\rceil}{d}
    }_{\text{VQ index cost}}
    +
    \underbrace{
        \frac{b_c Kd}
             {N}}_{\text{codebook cost}},
\end{equation}
where $N$ is the number of elements in the tensor and $b_c$ is the number of bits used to store each codebook element.

\begin{figure}[t]
    \centering
    \includegraphics[
        width=\columnwidth
    ]{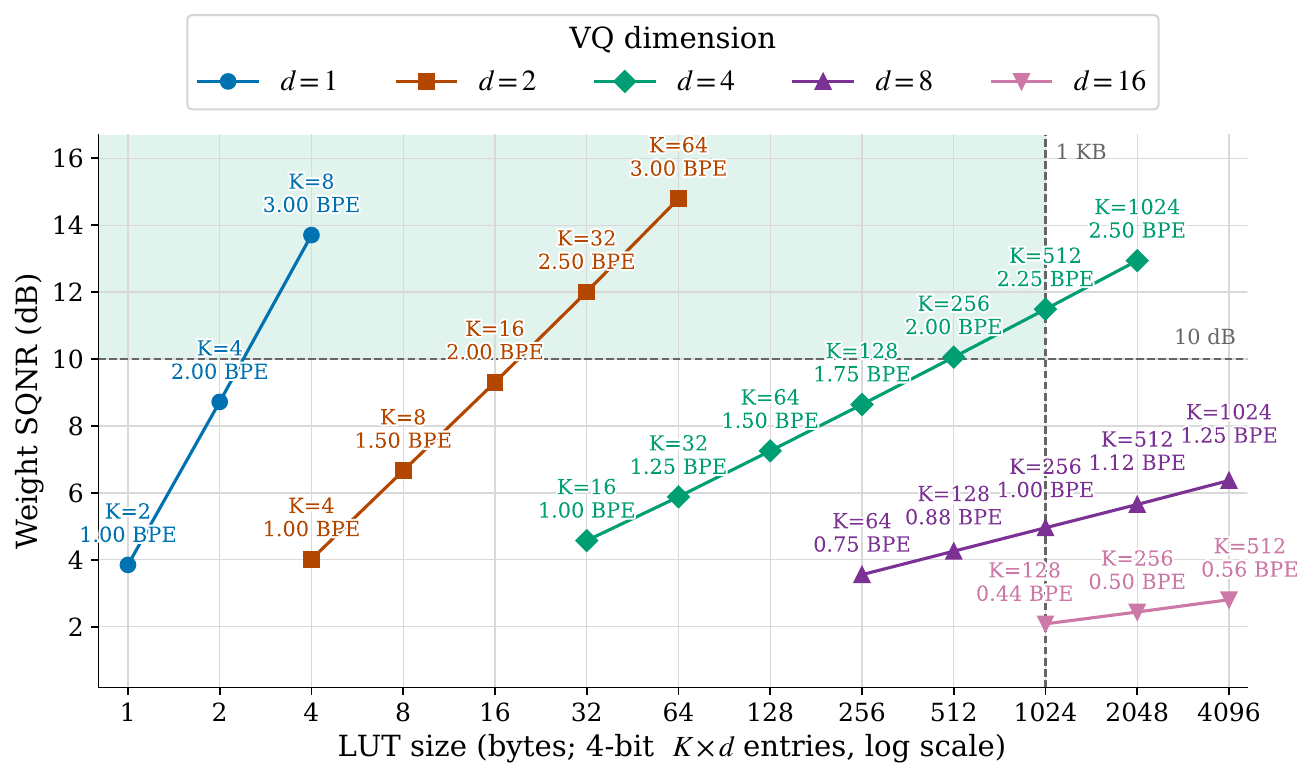}
    \caption{
        Aggregate weight SQNR as a function of effective codebook size for
        different VQ dimensions $d$. Each point corresponds to a different effective BPE and
        number of centroids $K$, indicated by its numeric label. BPE calculation only includes VQ indices. The shaded upper-left region denotes configurations that satisfy both design constraints: SQNR$\geq10$~dB and LUT size $\leq1$~KB.
    }
    \label{fig:sqlut}
\end{figure}

Figure~\ref{fig:sqlut} shows SQNR for various $d$ and $K$. As it can be seen in the Figure, larger VQ dimensions provide better representation quality at the same BPE by capturing correlations between neighboring weights.
However, larger dimensions also require more centroids, and therefore a larger look-up table (LUT) to achieve the same BPE to represent the higher-dimensional vector space.
Larger look-up tables cause higher storage and bandwidth needs and longer look-up read latencies.

Figure~\ref{fig:sqlut} shows that $d=4$ offers the best overall tradeoff, achieving the highest SQNR near the 2-bit regime among configurations with practical LUT sizes. Although $d \geq 8$ may achieve higher SQNR at the same BPE, doing so requires impractically large LUTs.
Empirically, we find that an SQNR of at least $10$~dB provides adequate model quality. For practicality, we also impose a $1$~KB LUT size limit with 4-bit codebook entries.
Although the appropriate SQNR threshold and the tradeoffs among SQNR, BPE, and LUT size depend on the evaluated model and the underlying hardware, we adopt these constraints as practical guidelines for our experiments.
In particular, $d=4$ with $K=256$ achieves an SQNR$\geq10$~dB at approximately $2$ BPE while keeping the LUT below $1$~KB. As demonstrated in later sections, this configuration preserves adequate end-to-end model accuracy. We therefore use $d=4$ and $K=256$ in the following experiments.

\begin{figure*}[t]
    \centering
    \includegraphics[
        width=\textwidth
    ]{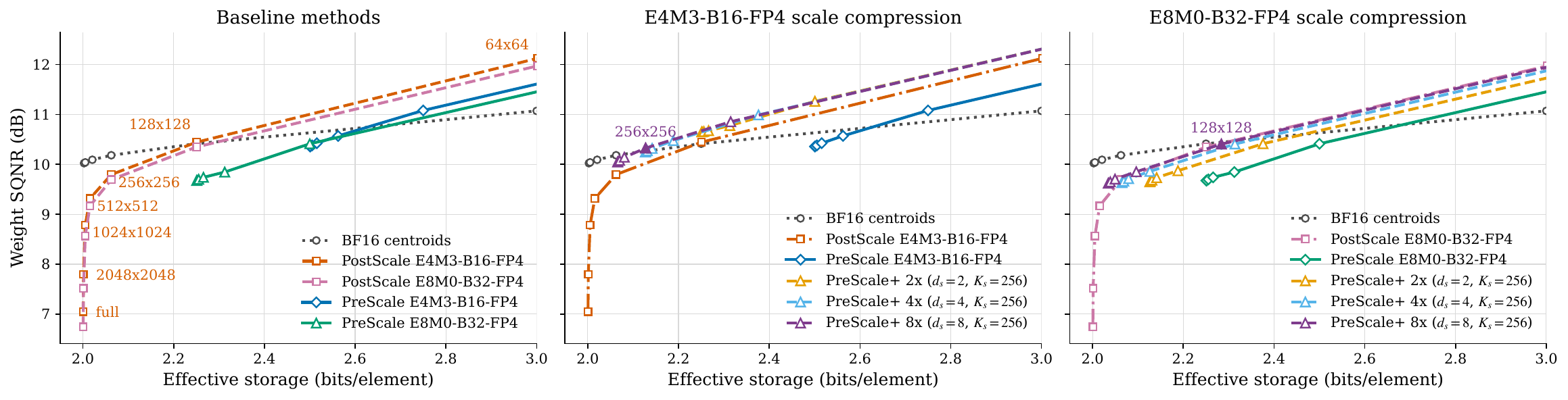}
    \caption{
        Ablation of FP4 codebook quantization. Each color shows the QBlock-size sweep from full tensor to $64\times 64$ for a VQ/MXCode scheme and datatype. The BF16-centroid baseline is independent of the MX scale format and is shown in gray across all three plots. From left to right, charts shows
        \textbf{(a)} QBlock-size sweep for E4M3-B16-FP4 and E8M0-B32-FP4.
        \textbf{(b)} QBlock-size sweep at various scale compression ratios for E4M3-B16-FP4 \mbox{MXCode-PreScale+}.
        \textbf{(c)} QBlock-size sweep at various scale compression ratios for E8M0-B32-FP4 \mbox{MXCode-PreScale+}.
        Effective storage includes the VQ indices, FP4 codebooks, MX scales,
        and scale-VQ metadata.
    }
    \label{fig:scbpe}
\end{figure*}

\subsubsection{\mbox{MXCode-PostScale}}

MXCode-PostScale directly quantizes  codebooks to MXFP4 formats, creating a substantially coarser scaling granularity than conventional MX quantization as described in Section~\ref{sec:post}.
Standard MX formats share a scale across a small block of elements, typically 16 or 32 values. In MXCode-PostScale, however, all $Kd$ scalar entries must share a single scale, for $K$ centroids with $d$ dimension. For example, for our configuration, $K{=}256$ and $d{=}4$, this requires sharing one scale across all 1024 scalar codebook entries.
Since every weight in the tensor is reconstructed from this codebook, the entire tensor is indirectly constrained by the same codebook scale.
While this approach has negligible scale overhead, a small number of large centroid values can consequently determine the scale and reduce the effective precision available to the remaining entries, causing a significant degradation in model quality.

To reduce this mismatch, we partition each weight tensor into smaller QBlocks of size $B_{\mathrm{out}}\times B_{\mathrm{in}}$ and learn an independent codebook for each QBlock as mentioned in Section~\ref{sec:pipeline}.
With smaller QBlocks, each codebook represents a more localized weight distribution and receives its own MX scale.
This reduces the dynamic range within each codebook and improves the accuracy of FP4 quantization.

Smaller QBlocks, however, increase the number of codebooks and therefore their amortized storage cost.
For PostScale FP4 codebooks, the effective storage cost is
\begin{equation}
    \mathrm{BPE}_{\mathrm{PostScale}}
    =
    \underbrace{
        \frac{\lceil \log_2 K\rceil}{d}
    }_{\text{VQ index cost}}
    +
    \underbrace{
        \frac{b_c Kd}
             {B_{\mathrm{out}}B_{\mathrm{in}}}
    }_{\text{codebook cost}}
    +
    \underbrace{
        \frac{b_s}
             {B_{\mathrm{out}}B_{\mathrm{in}}}
    }_{\text{scale cost}},
\end{equation}
where $b_c$ is the bit precision of each codebook element and $b_s$ is
the number of bits used to store one codebook scale. Smaller QBlocks increase codebook and scale costs, as each stores its own codebook and scale, imposing a minimum QBlock size for a given BPE budget.

In addition to their storage cost, smaller QBlocks increase the number
of codebooks that must be indexed, addressed, and managed by the
hardware. QBlock size therefore affects not only compression quality
and BPE, but also LUT organization and implementation
complexity.

\subsubsection{\mbox{MXCode-PreScale}}

Different from the PostScale approach, \mbox{MXCode-PreScale} extracts the MX-style scales first and perform VQ on the normalized weights.
This enables codebook to be learned directly in the normalized FP4 domain without requiring post-VQ scale estimation and avoids forcing all $Kd$ values in a codebook to share a scale. Consequently, \mbox{MXCode-PreScale} generally provides higher SQNR than post-VQ quantization at the same QBlock size. This improvement, however, comes with the cost of storing the extracted scales. If one $b_s$-bit scale is shared by every $S$ weight elements, the effective storage cost becomes
\begin{equation}
    \mathrm{BPE}_{\mathrm{PreScale}}
    =
    \underbrace{
        \frac{\lceil \log_2 K\rceil}{d}
    }_{\text{VQ index cost}}
    +
    \underbrace{
        \frac{b_cKd}
             {B_{\mathrm{out}}B_{\mathrm{in}}}
    }_{\text{codebook cost}}
    +
    \underbrace{
        \frac{b_s}{S}
    }_{\text{MX scales}}.
    \label{eq:prebpe}
\end{equation}
For an 8-bit scale, the scale overhead is $0.5$ BPE when $S=16$ and $0.25$ BPE when $S=32$. This overhead is independent of the QBlock size and is substantial in the approximately 2--3 BPE regime considered here.

Figure~\ref{fig:scbpe}(a) compares PostScale and PreScale approaches against VQ with BF16 codebook entries across various QBlock sizes from full-tensor size to $64\times 64$.

This plot shows that PreScale provides a head start in SQNR over PostScale at a fixed QBlock
size due to finer-granularity scaling, but shifts the corresponding operating points to a higher BPE.
Under a fixed storage budget, however, PostScale can instead spend
these bits on smaller QBlocks and more local codebooks.
As a result, PreScale does not necessarily provide a better
SQNR--BPE tradeoff when smaller QBlock sizes are available.

\subsubsection{\mbox{MXCode-PreScale+}}

\mbox{MXCode-PreScale+} retains the fine-grained scale extraction and compresses the
resulting scale tensor using VQ.
We group the extracted scales into
vectors of dimension $d_s$ and encode them using a scale codebook with
$K_s$ entries.

The resulting scale-storage overhead is
\begin{equation}
    \mathrm{BPE}_{\mathrm{scale\text{-}VQ}}
    =
    \underbrace{
        \frac{\lceil\log_2 K_s\rceil}{d_s S}
    }_{\text{scale-index cost}}
    +
    \underbrace{
        \frac{b_s K_s d_s}{N}
    }_{\text{scale-codebook cost}},
\end{equation}
where $N/S$ is the number of extracted scales. The
complete PreScale+ representation combines this overhead with the
element-index and element-codebook terms in
Eq.~\eqref{eq:prebpe}. Since one scale codebook is amortized over
a large weight tensor, scale storage cost can be reduced significantly compared to PreScale.

Panels (b) and (c) of Figure~\ref{fig:scbpe}
evaluate the effectiveness of \mbox{MXCode-PreScale+}.
We observe that MX scales have a
simpler distribution than the original weights and are therefore
highly compressible.
The plot illustrates that the scales can be compressed up to $8\times$, i.e., $d_s{=}8, K_s{=}256$  while largely preserving the SQNR of uncompressed
PreScale for both E4M3-B16-FP4 and E8M0-B32-FP4 datatypes.

\subsubsection{Design Trade-offs and Insights}
The relative storage--distortion trade-offs of PostScale,
PreScale, and PreScale+ depend on the scale configuration,
QBlock size, and target storage budget. Although uncompressed
PreScale can achieve higher SQNR at a fixed QBlock size,
its scale-storage overhead limits its competitiveness
under constrained BPE budgets. By reducing this overhead,
PreScale+ and PostScale provide additional operating points that
can offer a more favorable balance between storage and
reconstruction quality.

For E4M3-B16-FP4, PreScale+ generally achieves higher SQNR than
PostScale.
For E8M0-B32-FP4, the advantage of PreScale+ over
PostScale is smaller. In PreScale+, scales and normalized
values are compressed separately, and their errors interact
when recombined. Power-of-two scales shared across $32$
values provide coarser normalization than E4M3 scales
shared across $16$ values, potentially increasing this
reconstruction error, combined with extra scale compression. PostScale applies scaling after VQ,
avoiding the effect of coarse pre-normalization on VQ.

QBlock size introduces a further trade-off between
codebook overhead and local adaptation: smaller QBlocks
allow codebooks to capture more localized weight
distributions, but increase storage per element.
MXCode schemes generally approach the BF16-centroid
curve at $256\times256$ QBlocks for E4M3-B16-FP4 and
$128\times128$ QBlocks for E8M0-B32-FP4, with the
SQNR gap depending on the scheme. We therefore adopt these QBlock sizes across the MXCode schemes evaluated
subsequently. These choices provide representative low-storage operating
points, rather than universally optimal configurations, and
can be further tailored to the model's weight distribution,
downstream task, and hardware constraints.

For brevity, the main text does not report E5M3-B16-FP4
results. In our experiments, E4M3 block scales
combined with a high-precision tensor scale yield
similar precision to E5M3-B16-FP4 data format. E5M3-B16-FP4 and E4M3-B16-FP4 comparison is provided in
Appendix~\ref{app:e5m3}.

\begin{table*}[!t]
\centering
\caption{W4A4 evaluation across three models. Accuracy and standard errors are reported as percentages. Native FP4 uses W4A4 computation with either the E4M3-B16-FP4 or E8M0-B32-FP4 format; without any VQ or MXCode. Weight storage, measured in BPE, includes VQ indices, FP4 codebooks, and scales, as applicable. Average accuracy weights the ten benchmarks equally.}
\label{tab:accuracy}
\begingroup
\fontsize{8}{9}\selectfont
\setlength{\tabcolsep}{3pt}
\renewcommand{\arraystretch}{1.10}
\begin{tabular*}{\textwidth}{@{\extracolsep{\fill}}clcccccccc@{}}
\toprule
& \textbf{Compute format} & \multicolumn{4}{c}{\textbf{E4M3-B16-FP4}} & \multicolumn{4}{c}{\textbf{E8M0-B32-FP4}} \\
\cmidrule(lr){3-6} \cmidrule(lr){7-10}
& \textbf{Method} & \shortstack{\textbf{Native}\\\textbf{FP4}} & \shortstack{\textbf{MXCode}\\\textbf{PostScale}} & \shortstack{\textbf{MXCode}\\\textbf{PreScale}} & \shortstack{\textbf{MXCode}\\\textbf{PreScale+}} & \shortstack{\textbf{Native}\\\textbf{FP4}} & \shortstack{\textbf{MXCode}\\\textbf{PostScale}} & \shortstack{\textbf{MXCode}\\\textbf{PreScale}} & \shortstack{\textbf{MXCode}\\\textbf{PreScale+}} \\
\midrule
& \textbf{Weight (BPE)} & 4.50 & 2.06 & 2.56 & 2.13 & 4.25 & 2.25 & 2.50 & 2.28 \\
\midrule
\multirow{11}{*}{\rotatebox[origin=c]{90}{\textbf{Qwen3 (235B)}}} & MMLU-Pro & 82.41\,\textcolor{gray}{\(\pm 0.34\)} & 79.18\,\textcolor{gray}{\(\pm 0.36\)} & 80.34\,\textcolor{gray}{\(\pm 0.35\)} & 80.54\,\textcolor{gray}{\(\pm 0.35\)} & 82.02\,\textcolor{gray}{\(\pm 0.34\)} & 79.67\,\textcolor{gray}{\(\pm 0.35\)} & 79.09\,\textcolor{gray}{\(\pm 0.36\)} & 79.00\,\textcolor{gray}{\(\pm 0.36\)} \\
 & RULER & 95.05\,\textcolor{gray}{\(\pm 0.14\)} & 94.16\,\textcolor{gray}{\(\pm 0.15\)} & 95.01\,\textcolor{gray}{\(\pm 0.14\)} & 95.00\,\textcolor{gray}{\(\pm 0.14\)} & 95.13\,\textcolor{gray}{\(\pm 0.13\)} & 94.69\,\textcolor{gray}{\(\pm 0.14\)} & 94.73\,\textcolor{gray}{\(\pm 0.14\)} & 94.83\,\textcolor{gray}{\(\pm 0.14\)} \\
 & DROP & 88.56\,\textcolor{gray}{\(\pm 0.33\)} & 86.98\,\textcolor{gray}{\(\pm 0.34\)} & 86.54\,\textcolor{gray}{\(\pm 0.35\)} & 87.69\,\textcolor{gray}{\(\pm 0.34\)} & 87.96\,\textcolor{gray}{\(\pm 0.33\)} & 85.55\,\textcolor{gray}{\(\pm 0.36\)} & 87.52\,\textcolor{gray}{\(\pm 0.34\)} & 86.35\,\textcolor{gray}{\(\pm 0.35\)} \\
 & GSM8K & 96.89\,\textcolor{gray}{\(\pm 0.48\)} & 96.66\,\textcolor{gray}{\(\pm 0.49\)} & 97.12\,\textcolor{gray}{\(\pm 0.46\)} & 96.36\,\textcolor{gray}{\(\pm 0.52\)} & 96.89\,\textcolor{gray}{\(\pm 0.48\)} & 96.82\,\textcolor{gray}{\(\pm 0.48\)} & 96.44\,\textcolor{gray}{\(\pm 0.51\)} & 96.21\,\textcolor{gray}{\(\pm 0.53\)} \\
 & IFEval & 84.47\,\textcolor{gray}{\(\pm 1.56\)} & 79.48\,\textcolor{gray}{\(\pm 1.74\)} & 81.70\,\textcolor{gray}{\(\pm 1.66\)} & 81.33\,\textcolor{gray}{\(\pm 1.68\)} & 84.29\,\textcolor{gray}{\(\pm 1.57\)} & 81.15\,\textcolor{gray}{\(\pm 1.68\)} & 81.89\,\textcolor{gray}{\(\pm 1.66\)} & 82.62\,\textcolor{gray}{\(\pm 1.63\)} \\
 & HumanEval & 96.95\,\textcolor{gray}{\(\pm 1.35\)} & 90.24\,\textcolor{gray}{\(\pm 2.32\)} & 98.17\,\textcolor{gray}{\(\pm 1.05\)} & 95.73\,\textcolor{gray}{\(\pm 1.58\)} & 96.95\,\textcolor{gray}{\(\pm 1.35\)} & 96.95\,\textcolor{gray}{\(\pm 1.35\)} & 93.90\,\textcolor{gray}{\(\pm 1.87\)} & 95.73\,\textcolor{gray}{\(\pm 1.58\)} \\
 & MATH-500 & 97.00\,\textcolor{gray}{\(\pm 0.76\)} & 95.20\,\textcolor{gray}{\(\pm 0.96\)} & 95.60\,\textcolor{gray}{\(\pm 0.92\)} & 96.00\,\textcolor{gray}{\(\pm 0.88\)} & 96.80\,\textcolor{gray}{\(\pm 0.79\)} & 95.60\,\textcolor{gray}{\(\pm 0.92\)} & 95.60\,\textcolor{gray}{\(\pm 0.92\)} & 96.00\,\textcolor{gray}{\(\pm 0.88\)} \\
 & GPQA-Diamond & 67.17\,\textcolor{gray}{\(\pm 3.35\)} & 64.65\,\textcolor{gray}{\(\pm 3.41\)} & 64.14\,\textcolor{gray}{\(\pm 3.42\)} & 64.14\,\textcolor{gray}{\(\pm 3.42\)} & 68.69\,\textcolor{gray}{\(\pm 3.30\)} & 67.68\,\textcolor{gray}{\(\pm 3.33\)} & 67.17\,\textcolor{gray}{\(\pm 3.35\)} & 67.68\,\textcolor{gray}{\(\pm 3.33\)} \\
 & MGSM & 86.51\,\textcolor{gray}{\(\pm 0.63\)} & 90.69\,\textcolor{gray}{\(\pm 0.55\)} & 91.16\,\textcolor{gray}{\(\pm 0.54\)} & 90.15\,\textcolor{gray}{\(\pm 0.56\)} & 86.62\,\textcolor{gray}{\(\pm 0.63\)} & 88.25\,\textcolor{gray}{\(\pm 0.60\)} & 90.22\,\textcolor{gray}{\(\pm 0.56\)} & 86.25\,\textcolor{gray}{\(\pm 0.64\)} \\
 & MBPP & 94.40\,\textcolor{gray}{\(\pm 1.03\)} & 89.40\,\textcolor{gray}{\(\pm 1.38\)} & 91.80\,\textcolor{gray}{\(\pm 1.23\)} & 91.40\,\textcolor{gray}{\(\pm 1.26\)} & 93.00\,\textcolor{gray}{\(\pm 1.14\)} & 91.80\,\textcolor{gray}{\(\pm 1.23\)} & 92.20\,\textcolor{gray}{\(\pm 1.20\)} & 91.60\,\textcolor{gray}{\(\pm 1.24\)} \\
 & \textbf{Average accuracy} & 88.94 & 86.66 & 88.16 & 87.83 & 88.84 & 87.82 & 87.88 & 87.63 \\
\midrule
\multirow{11}{*}{\rotatebox[origin=c]{90}{\textbf{DeepSeek-R1 (671B)}}} & MMLU-Pro & 84.23\,\textcolor{gray}{\(\pm 0.33\)} & 82.86\,\textcolor{gray}{\(\pm 0.34\)} & 83.35\,\textcolor{gray}{\(\pm 0.33\)} & 82.94\,\textcolor{gray}{\(\pm 0.34\)} & 83.61\,\textcolor{gray}{\(\pm 0.33\)} & 81.70\,\textcolor{gray}{\(\pm 0.34\)} & 82.33\,\textcolor{gray}{\(\pm 0.34\)} & 82.52\,\textcolor{gray}{\(\pm 0.34\)} \\
 & RULER & 94.42\,\textcolor{gray}{\(\pm 0.14\)} & 94.00\,\textcolor{gray}{\(\pm 0.15\)} & 94.16\,\textcolor{gray}{\(\pm 0.15\)} & 94.53\,\textcolor{gray}{\(\pm 0.14\)} & 94.05\,\textcolor{gray}{\(\pm 0.15\)} & 93.06\,\textcolor{gray}{\(\pm 0.16\)} & 93.84\,\textcolor{gray}{\(\pm 0.15\)} & 93.84\,\textcolor{gray}{\(\pm 0.15\)} \\
 & DROP & 88.34\,\textcolor{gray}{\(\pm 0.33\)} & 87.78\,\textcolor{gray}{\(\pm 0.34\)} & 88.54\,\textcolor{gray}{\(\pm 0.33\)} & 88.51\,\textcolor{gray}{\(\pm 0.33\)} & 88.13\,\textcolor{gray}{\(\pm 0.33\)} & 87.79\,\textcolor{gray}{\(\pm 0.34\)} & 88.64\,\textcolor{gray}{\(\pm 0.32\)} & 88.23\,\textcolor{gray}{\(\pm 0.33\)} \\
 & GSM8K & 96.59\,\textcolor{gray}{\(\pm 0.50\)} & 96.29\,\textcolor{gray}{\(\pm 0.52\)} & 96.74\,\textcolor{gray}{\(\pm 0.49\)} & 96.66\,\textcolor{gray}{\(\pm 0.49\)} & 96.97\,\textcolor{gray}{\(\pm 0.47\)} & 96.21\,\textcolor{gray}{\(\pm 0.53\)} & 96.66\,\textcolor{gray}{\(\pm 0.49\)} & 96.51\,\textcolor{gray}{\(\pm 0.51\)} \\
 & IFEval & 85.03\,\textcolor{gray}{\(\pm 1.54\)} & 83.92\,\textcolor{gray}{\(\pm 1.58\)} & 83.73\,\textcolor{gray}{\(\pm 1.59\)} & 81.70\,\textcolor{gray}{\(\pm 1.66\)} & 83.73\,\textcolor{gray}{\(\pm 1.59\)} & 81.33\,\textcolor{gray}{\(\pm 1.68\)} & 82.44\,\textcolor{gray}{\(\pm 1.64\)} & 84.47\,\textcolor{gray}{\(\pm 1.56\)} \\
 & HumanEval & 97.56\,\textcolor{gray}{\(\pm 1.21\)} & 96.95\,\textcolor{gray}{\(\pm 1.35\)} & 96.95\,\textcolor{gray}{\(\pm 1.35\)} & 97.56\,\textcolor{gray}{\(\pm 1.21\)} & 98.17\,\textcolor{gray}{\(\pm 1.05\)} & 93.90\,\textcolor{gray}{\(\pm 1.87\)} & 94.51\,\textcolor{gray}{\(\pm 1.78\)} & 95.73\,\textcolor{gray}{\(\pm 1.58\)} \\
 & MATH-500 & 96.80\,\textcolor{gray}{\(\pm 0.79\)} & 96.20\,\textcolor{gray}{\(\pm 0.86\)} & 96.20\,\textcolor{gray}{\(\pm 0.86\)} & 95.80\,\textcolor{gray}{\(\pm 0.90\)} & 96.20\,\textcolor{gray}{\(\pm 0.86\)} & 93.60\,\textcolor{gray}{\(\pm 1.10\)} & 95.20\,\textcolor{gray}{\(\pm 0.96\)} & 96.20\,\textcolor{gray}{\(\pm 0.86\)} \\
 & GPQA-Diamond & 68.18\,\textcolor{gray}{\(\pm 3.32\)} & 69.70\,\textcolor{gray}{\(\pm 3.27\)} & 67.68\,\textcolor{gray}{\(\pm 3.33\)} & 69.70\,\textcolor{gray}{\(\pm 3.27\)} & 69.19\,\textcolor{gray}{\(\pm 3.29\)} & 63.13\,\textcolor{gray}{\(\pm 3.44\)} & 67.68\,\textcolor{gray}{\(\pm 3.33\)} & 65.66\,\textcolor{gray}{\(\pm 3.38\)} \\
 & MGSM & 91.45\,\textcolor{gray}{\(\pm 0.53\)} & 92.36\,\textcolor{gray}{\(\pm 0.50\)} & 92.33\,\textcolor{gray}{\(\pm 0.51\)} & 91.05\,\textcolor{gray}{\(\pm 0.54\)} & 91.45\,\textcolor{gray}{\(\pm 0.53\)} & 91.02\,\textcolor{gray}{\(\pm 0.54\)} & 92.22\,\textcolor{gray}{\(\pm 0.51\)} & 91.38\,\textcolor{gray}{\(\pm 0.53\)} \\
 & MBPP & 93.40\,\textcolor{gray}{\(\pm 1.11\)} & 93.40\,\textcolor{gray}{\(\pm 1.11\)} & 93.20\,\textcolor{gray}{\(\pm 1.13\)} & 93.40\,\textcolor{gray}{\(\pm 1.11\)} & 92.60\,\textcolor{gray}{\(\pm 1.17\)} & 91.00\,\textcolor{gray}{\(\pm 1.28\)} & 93.40\,\textcolor{gray}{\(\pm 1.11\)} & 92.40\,\textcolor{gray}{\(\pm 1.19\)} \\
 & \textbf{Average accuracy} & 89.60 & 89.35 & 89.29 & 89.19 & 89.41 & 87.27 & 88.69 & 88.69 \\
\midrule
\multirow{11}{*}{\rotatebox[origin=c]{90}{\textbf{GLM-5.2 (753B)}}} & MMLU-Pro & 87.21\,\textcolor{gray}{\(\pm 0.30\)} & 85.46\,\textcolor{gray}{\(\pm 0.31\)} & 86.29\,\textcolor{gray}{\(\pm 0.31\)} & 84.50\,\textcolor{gray}{\(\pm 0.32\)} & 87.08\,\textcolor{gray}{\(\pm 0.30\)} & 85.24\,\textcolor{gray}{\(\pm 0.32\)} & 85.27\,\textcolor{gray}{\(\pm 0.32\)} & 82.58\,\textcolor{gray}{\(\pm 0.34\)} \\
 & RULER & 95.71\,\textcolor{gray}{\(\pm 0.13\)} & 96.22\,\textcolor{gray}{\(\pm 0.12\)} & 96.73\,\textcolor{gray}{\(\pm 0.12\)} & 95.67\,\textcolor{gray}{\(\pm 0.13\)} & 95.85\,\textcolor{gray}{\(\pm 0.13\)} & 95.38\,\textcolor{gray}{\(\pm 0.13\)} & 95.79\,\textcolor{gray}{\(\pm 0.13\)} & 91.16\,\textcolor{gray}{\(\pm 0.17\)} \\
 & DROP & 91.98\,\textcolor{gray}{\(\pm 0.28\)} & 91.23\,\textcolor{gray}{\(\pm 0.29\)} & 90.80\,\textcolor{gray}{\(\pm 0.30\)} & 90.06\,\textcolor{gray}{\(\pm 0.31\)} & 91.78\,\textcolor{gray}{\(\pm 0.28\)} & 90.91\,\textcolor{gray}{\(\pm 0.29\)} & 89.67\,\textcolor{gray}{\(\pm 0.31\)} & 89.18\,\textcolor{gray}{\(\pm 0.32\)} \\
 & GSM8K & 97.50\,\textcolor{gray}{\(\pm 0.43\)} & 96.74\,\textcolor{gray}{\(\pm 0.49\)} & 96.89\,\textcolor{gray}{\(\pm 0.48\)} & 96.82\,\textcolor{gray}{\(\pm 0.48\)} & 97.57\,\textcolor{gray}{\(\pm 0.42\)} & 96.89\,\textcolor{gray}{\(\pm 0.48\)} & 96.97\,\textcolor{gray}{\(\pm 0.47\)} & 96.36\,\textcolor{gray}{\(\pm 0.52\)} \\
 & IFEval & 94.09\,\textcolor{gray}{\(\pm 1.02\)} & 93.16\,\textcolor{gray}{\(\pm 1.09\)} & 92.05\,\textcolor{gray}{\(\pm 1.16\)} & 89.46\,\textcolor{gray}{\(\pm 1.32\)} & 93.53\,\textcolor{gray}{\(\pm 1.06\)} & 91.68\,\textcolor{gray}{\(\pm 1.19\)} & 92.61\,\textcolor{gray}{\(\pm 1.19\)} & 90.94\,\textcolor{gray}{\(\pm 1.24\)} \\
 & HumanEval & 99.39\,\textcolor{gray}{\(\pm 0.61\)} & 97.56\,\textcolor{gray}{\(\pm 1.21\)} & 99.39\,\textcolor{gray}{\(\pm 0.61\)} & 98.78\,\textcolor{gray}{\(\pm 0.86\)} & 100.00\,\textcolor{gray}{\(\pm 0.00\)} & 98.17\,\textcolor{gray}{\(\pm 1.05\)} & 97.56\,\textcolor{gray}{\(\pm 1.21\)} & 97.56\,\textcolor{gray}{\(\pm 1.21\)} \\
 & MATH-500 & 98.00\,\textcolor{gray}{\(\pm 0.63\)} & 97.80\,\textcolor{gray}{\(\pm 0.66\)} & 97.60\,\textcolor{gray}{\(\pm 0.69\)} & 97.40\,\textcolor{gray}{\(\pm 0.71\)} & 97.80\,\textcolor{gray}{\(\pm 0.66\)} & 97.60\,\textcolor{gray}{\(\pm 0.69\)} & 96.20\,\textcolor{gray}{\(\pm 0.86\)} & 96.40\,\textcolor{gray}{\(\pm 0.83\)} \\
 & GPQA-Diamond & 82.83\,\textcolor{gray}{\(\pm 2.69\)} & 83.33\,\textcolor{gray}{\(\pm 2.66\)} & 84.85\,\textcolor{gray}{\(\pm 2.55\)} & 85.35\,\textcolor{gray}{\(\pm 2.52\)} & 87.88\,\textcolor{gray}{\(\pm 2.33\)} & 87.37\,\textcolor{gray}{\(\pm 2.37\)} & 85.35\,\textcolor{gray}{\(\pm 2.51\)} & 79.80\,\textcolor{gray}{\(\pm 2.86\)} \\
 & MGSM & 94.25\,\textcolor{gray}{\(\pm 0.44\)} & 93.27\,\textcolor{gray}{\(\pm 0.48\)} & 93.56\,\textcolor{gray}{\(\pm 0.47\)} & 92.69\,\textcolor{gray}{\(\pm 0.49\)} & 94.47\,\textcolor{gray}{\(\pm 0.44\)} & 93.27\,\textcolor{gray}{\(\pm 0.48\)} & 94.86\,\textcolor{gray}{\(\pm 0.48\)} & 92.51\,\textcolor{gray}{\(\pm 0.50\)} \\
 & MBPP & 96.60\,\textcolor{gray}{\(\pm 0.81\)} & 94.80\,\textcolor{gray}{\(\pm 0.99\)} & 95.40\,\textcolor{gray}{\(\pm 0.94\)} & 94.40\,\textcolor{gray}{\(\pm 1.03\)} & 95.80\,\textcolor{gray}{\(\pm 0.90\)} & 95.00\,\textcolor{gray}{\(\pm 0.98\)} & 92.80\,\textcolor{gray}{\(\pm 1.16\)} & 94.80\,\textcolor{gray}{\(\pm 0.99\)} \\
 & \textbf{Average accuracy} & 93.76 & 92.96 & 93.36 & 92.51 & 94.18 & 93.15 & 92.71 & 91.13 \\
\bottomrule
\end{tabular*}
\endgroup
\end{table*}

\subsection{Accuracy and Benchmarks}
\label{sec:accuracy}

Table~\ref{tab:accuracy} compares Native FP4 and MXCode quantization methods for Qwen3 (235B) \citep{qwen25}, DeepSeek-R1 (671B) \citep{dsr125}, and GLM-5.2 (753B) \citep{glm526} across ten benchmarks spanning knowledge, reasoning, instruction following, code generation, mathematics, and long-context understanding. We evaluate model quality using the Language Model Evaluation Harness (lm-eval-harness) \citep{lmeval23}, with inference served through SGLang and MX-compatible numerics emulated using the Microsoft MX Emulation Library. We use benchmark task definitions and each model’s native chat template, preserving task-specific instructions, answer-extraction rules, and few-shot examples.
All E4M3-B16-FP4 configurations use $256\times256$ QBlocks, whereas the E8M0-B32-FP4 configurations use $128\times128$ QBlocks, following the ablation results in Section~\ref{sec:ablation} PreScale+ uses $8\times$ compression for scales compared to PreScale.

Overall, all three MXCode schemes achieve competitive accuracy relative to native FP4 while substantially reducing weight storage. Individual benchmarks, however, highlight complementary strengths and variations across the three schemes. This variability is particularly evident for Qwen3 (235B), the smallest model evaluated. Under E4M3-B16-FP4, for example, HumanEval accuracy ranges from 90.24\% to 98.17\% across MXCode schemes, compared with 96.95\% for native FP4. MXCode also improves MGSM accuracy for Qwen3 (235B), despite more aggressive compression, while tasks such as IFEval exhibit some degradation in certain cases.

PreScale often provides stronger accuracy by retaining uncompressed scales,
whereas PostScale and PreScale+ offer lower-storage operating points.
This advantage is not universal, however: the preferred method depends
on both the model and the compute format.

For Qwen3 (235B), PreScale achieves the highest MXCode average accuracy
for both E4M3-B16-FP4 and E8M0-B32-FP4.
Under E4M3-B16-FP4, PreScale+ performs better than PostScale, providing
a lower-storage alternative to PreScale at 2.13 BPE.
Under E8M0-B32-FP4, PreScale reaches 87.88\% at 2.50 BPE, but PostScale is
nearly as accurate at 87.82\% while requiring only 2.25 BPE.

For DeepSeek-R1 (671B), PostScale provides the strongest E4M3-B16-FP4 trade-off:
89.35\% average accuracy at 2.06 BPE, only 0.25 percentage points
below native FP4.
Under E8M0-B32-FP4, PreScale and PreScale+ both achieve 88.69\%;
PreScale+ is preferable on storage, requiring 2.28 rather than
2.50 BPE. Both outperform PostScale's 87.27\%.

For GLM-5.2 (753B), PreScale provides
the highest E4M3-B16-FP4 MXCode average, with an estimated 93.36\% at
2.56 BPE, versus 93.76\% for native FP4. PostScale offers 92.96\% at the lower storage of 2.06 BPE.
Under E8M0-B32-FP4, PostScale achieves the highest MXCode average,
93.15\% at 2.25 BPE, exceeding both PreScale variants.

Overall, these results reveal a workload-dependent storage–quality trade-off: MXCode substantially reduces weight storage, but its quality impact varies across models and benchmarks, with instruction-following and code-generation tasks showing particular sensitivity and underscoring the importance of workload-specific evaluation.

\subsection{Performance Evaluation}

In this section, we examine whether MXCode's ability to enable MXFP4 compute along with compact representation translates
into practical GPU serving benefits.

\subsubsection{Experimental Setup}

We evaluate Qwen3 (235B) on a single node consisting of 4 - NVIDIA GB300 GPUs using SGLang
and an InferenceX-style streaming benchmark client. We use the same checkpoint as in the accuracy evaluation in Table \ref{tab:accuracy} to run experiments in this section. All performance
experiments consistently make use of requests with 32K input tokens and 1K generated tokens unless otherwise stated. We choose 32K-input/1K-output requests to combine substantial KV-cache demand with long-context prefill and sustained decoding, exposing both the memory-capacity and compute-efficiency tradeoffs of MXCode.
Our kernels target the MoE feed-forward layers, including the
gate/up and down projections. Non-expert parameters, activations
at the layer interfaces, and the KV cache remain in BF16 for this section. We discuss three serving implementations for comparison.

\paragraph{Vanilla VQ (BF16): Compressed storage with BF16 compute.}
We include a high-precision-codebook VQ reference to represent
conventional VQ inference. Expert weights are stored as compact
integer indices along with BF16 codebooks. This implementation reconstructs BF16 weights on the GPU for matrix multiplications, without retaining expanded weights across
invocations. This version uses fused VQ execution and codebook caching for best performance.
We use the same $\sim$2BPE configuration i.e., VQ dim, Qblock size, and codebook size used in MXCode for fairness.

\paragraph{Native FP4: Explicit FP4 weights with FP4 compute.}
Native FP4 provides the
complementary reference for FP4 computation without VQ compression,
using resident E4M3-B16-FP4 weights and FlashInfer/CUTLASS fused MoE.
We do not report a vendor E8M0-B32-FP4 baseline because we did
not identify a compatible off-the-shelf MoE kernel in the
evaluated software stack.

Together, Vanilla VQ (BF16) and Native FP4 provide complementary
baselines: the former combines VQ-compressed storage with BF16
computation, while the latter uses explicit FP4 weights with native
FP4 computation. These references help contextualize MXCode's
combination of VQ compression and FP4 execution.

\paragraph{MXCode: compressed storage with FP4 computation.}
MXCode brings native FP4 Tensor Core computation into the VQ
regime by combining compact integer indices with FP4 codebooks.
We evaluate MXCode (E4M3-B16-FP4) and
MXCode (E8M0-B32-FP4), using checkpoint evaluated in Table~\ref{tab:accuracy} under the
quantization configurations described in Section~\ref{sec:accuracy}.
Although MXCode-PreScale and
MXCode-PostScale differ in how codebooks and scales are obtained,
both ultimately provide FP4 operands and block scales for FP4 matmul. They can
therefore reuse the same FP4 projection kernel after
representation-specific operand preparation. All MXCode results
in this section use this shared FP4 projection backend. MXCode-PreScale+ kernel is excluded from the scope for brevity. We also exclude E5M3-B16-FP4 as there is no current hardware support for this datatype.

\begin{figure}[t]
  \centering
    \includegraphics[width=\columnwidth]{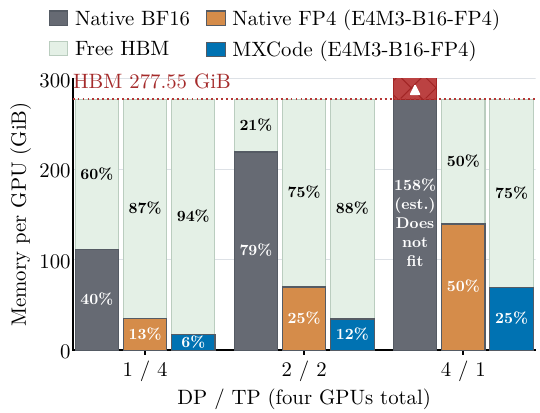}
  \caption{Per-GPU model-load memory for Qwen3 (235B).
MXCode denotes the shared
Pre/PostScale FP4 projection backend; the plotted footprint
uses PostScale-style codebook. Percentages in black indicate free HBM that can be used by KV-cache. MXCode (E8M0-B32-FP4) \& Vanilla VQ (BF16) has similiar footprint to MXCode. \vspace{-20pt}}
  \label{fig:mem}
\end{figure}

\subsubsection{Memory Footprint and Deployment Flexibility}

We first examine how expert-weight compression changes the memory
available for serving. Figure~\ref{fig:mem} compares
the per-GPU model-load footprint of Qwen3 (235B) across native
BF16, Native FP4 (E4M3-B16-FP4), and MXCode (E4M3-B16-FP4),
keeping the hardware budget fixed at four NVIDIA GB300 GPUs.
Relative to Native FP4, MXCode reduces the model-load footprint
by up to 2$\times$
and up to 6.6$\times$ when compared to Native BF16 across the three sharding schemes,
leaving more
memory available for serving state. MXCode (E8M0-B32-FP4) and Vanilla VQ (BF16), both at approximately 2 BPE, exhibit a similar memory-footprint trend to MXCode (E4M3-B16-FP4).

When comparing VQ solutions to Native FP4, a deployment can retain its existing DP/TP configuration and use
this additional headroom for KV capacity. Alternatively, the storage savings can accommodate a lower TP
width, reducing the number of GPUs required per replica while
preserving comparable memory headroom. In the figure,
MXCode by doubling DP and reducing TP at 2/2 has nearly the same remaining HBM per GPU compared to
Native FP4 DP/TP=1/4; similarly, MXCode at
4/1 is comparable to Native FP4 at 2/2.
Here, the total GPU budget remains four, so using fewer GPUs per
replica is illustrated by doubling DP rather than reducing the
total system size.

These comparisons demonstrate additional deployment choices,
not a preference for one sharding scheme over another.
Having established the memory-saving benefits of VQ over native FP4, we next evaluate its inference performance. Unless otherwise specified, all subsequent experiments use DP/TP = 1/1 (Single GPU) to isolate the effects of sharding.

\subsubsection{Serving Throughput}

We evaluate serving throughput and P90 interactivity for
32K-input/1K-output requests at client concurrencies
$C\in\{1,4,8,16,32\}$ using single GPU.
For each FP4 configuration, we report five benchmark repetitions,
each measuring $\max(32,2C)$ requests after excluded
workload-warmup requests sufficient to reach steady-state execution. Prefix caching is disabled, and the
static-HBM allocation fraction is 95\%. We report request-level P90 metrics with 95\% confidence intervals.

\begin{figure}[t]
  \centering
  \includegraphics[width=\columnwidth]{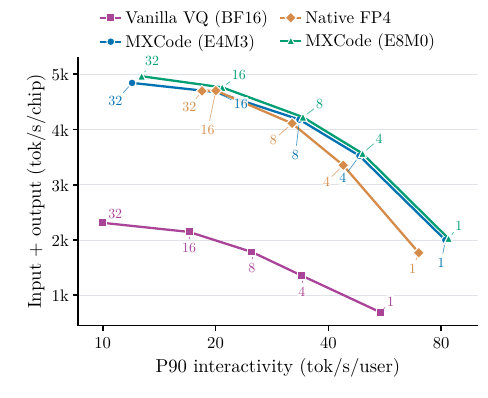}
  \caption{Serving throughput vs. P90 interactivity for 32K-input/1K-output
requests on Qwen3 (235B) using single GB300 GPU.
Point labels indicate client concurrency, including
queued requests. MXCode denotes the shared Pre/PostScale FP4 projection backend.\vspace{-10pt}}
  \label{fig:tput}
\end{figure}

Figure~\ref{fig:tput} shows that MXCode retains
near-Native-FP4 serving throughput with compact indexed weights,
including software lookup and reconstruction costs.
Client concurrency counts both running and queued requests, so the
curves reflect GPU execution and available KV capacity.
At C1, MXCode-E4M3 and MXCode-E8M0 achieve 2.00k and 2.03k input-plus-output tokens/s/chip versus 1.77k for Native FP4, corresponding to 13.0\%
and 14.5\% higher throughput, respectively, with P90
interactivity of 82.1 and 83.7 versus 69.7 tokens/s/user. With one active request and no admission queue, this reflects faster
decoding.
Both throughput and P90 interactivity improve at C4 and C8.

Native FP4 approaches throughput saturation at C=16, while MXCode accommodates additional concurrent requests through its reduced weight-memory footprint. At C=32, the allocated BF16 KV pools accommodate 31 complete 32K/1K contexts for MXCode versus 20 for Native FP4. MXCode-E4M3 and MXCode-E8M0 achieve 4.84k and 4.96k input-plus-output tokens/s/chip, respectively, compared with 4.70k for Native FP4. The additional HBM capacity enables greater resident concurrency or more reusable context storage, allowing deployments to balance throughput, latency, and context capacity.

Our serving measurements show that, at C1, MXCode-E4M3 and MXCode-E8M0 reduce mean TPOT by 15.1\% and 16.7\%, respectively, compared with Native FP4, while maintaining TTFT within 1\% of Native FP4. Further investigation shows that MXCode-E4M3 and Native FP4 achieve comparable prefill execution times at both C1 and C32, with differences below 1\% under both workloads.

At C32, MXCode-E4M3 achieves lower TTFT, primarily due to reduced scheduler waiting time. However, its per-user TPOT is higher because of slower token-generation steps with a larger active batch, as well as pauses caused by the GPU processing prompt chunks from other requests. Thus, increasing the number of concurrent generations can improve aggregate throughput, even when individual users experience longer token intervals.

Our prefill path reconstructs active FP4 weights for reuse with
grouped GEMM, whereas decode reconstructs tiles inside projection.
Separate profiling attributes 20.4\% of the MXCode prefill
expert-layer GPU timeline to explicit reconstruction.
Indexed-FP4 hardware could target this preparation cost while
preserving compressed storage.
Ideally eliminating the explicit reconstruction time, with all
other costs fixed, leaves 79.6\% of MXCode's current prefill
expert-layer time, giving an idealized speedup of
$1/0.796\approx1.26\times$; These measurements characterize the current software kernels, not
the performance limit of indexed-FP4 execution. Hardware/software co-design of codebook access, tile reconstruction,
and tensor-core operand delivery could reduce preparation overhead
and temporary expanded-weight traffic while preserving compact storage.
This motivates native indexed-FP4 support as a direction for further
acceleration; end-to-end hardware gains remain to be evaluated.

\subsubsection{Memory-Driven Gains with Prefix Reuse}

When native FP4 computation is already available, MXCode adds compression at the weight-storage level. Representing expert weights with compact indices and FP4 codebooks leaves more HBM for KV cache, supporting more simultaneous requests or greater reusable-context retention. We examine how these memory savings translate into serving benefits while accounting for software lookup and reconstruction costs.
We therefore examine a
prefix-reuse workload in which the expert-weight footprint affects
whether reusable context can remain resident. Here, additional
compression can improve performance by avoiding recomputation.

Each request contains a 32,512-token document prefix and a unique
256-token query making it 32,768-token prompt, followed by 1K generated tokens. Requests reuse
a shared set of 28 documents, with 4 active requests throughout.
All methods use single GPU, BF16 KV, same prefix-cache policy,
and 97.5\% static-HBM setting. Each repetition measures
512 queries after two cache-warmup sweeps.

\begin{table}[t]
  \centering
  \caption{32K/1K prefix-reuse latency  using one GB300 per method on Qwen3 (235B) model. Values are means of per-run 90th percentiles after post-warmup repetitions; parentheses
    show latency reductions relative to Native FP4. MXCode denotes the shared
PreScale/PostScale FP4 projection backend.}
  \label{tab:reuse}
  \begingroup
  \fontsize{8}{9}\selectfont
  \setlength{\tabcolsep}{3pt}
  \renewcommand{\arraystretch}{1.10}
  \begin{tabular*}{\columnwidth}{@{\extracolsep{\fill}}lccc@{}}
    \toprule
    Method & \shortstack{P90 TTFT\\(s)}
           & \shortstack{P90 TPOT\\(ms)}
           & \shortstack{P90 E2E\\(s)} \\
    \midrule
    \shortstack[l]{Native FP4\\E4M3-B16-FP4}
      & 8.688 & 27.24 & 32.52 \\
    \addlinespace
    \shortstack[l]{MXCode\\E4M3-B16-FP4}
      & \shortstack{0.486\\\textbf{(-94.4\%)}}
      & \shortstack{20.99\\\textbf{(-23.0\%)}}
      & \shortstack{21.87\\\textbf{(-32.8\%)}} \\
    \addlinespace
    \shortstack[l]{MXCode\\E8M0-B32-FP4}
      & \shortstack{0.477\\\textbf{(-94.5\%)}}
      & \shortstack{20.56\\\textbf{(-24.5\%)}}
      & \shortstack{21.43\\\textbf{(-34.1\%)}} \\
    \bottomrule
  \end{tabular*}
  \endgroup
\end{table}

MXCode's smaller expert footprint allows the runtime to allocate
approximately 1.11 million KV token slots, compared with
0.72 million for Native FP4. Consequently, MXCode achieves a
100\% complete-prefix hit rate, whereas the Native FP4
achieves 73.4\%. The Native FP4 kernel must evict cached prefixes and repeat
prefill work; MXCode retains the working set and processes only
the unique query suffix on each cache hit. In this experiment, average recomputed
prefill tokens fall from 7,580 to 256 per request for MXCode.

Table ~\ref{tab:reuse} shows the responsiveness gains from improved residency. Across both MXCode formats, P90 TTFT, TPOT, and end-to-end latency decrease by 94.4--94.5\%, 23.0--24.5\%, and 32.8--34.1\%, respectively.

\section{Lessons Learned and Implications}

MXCode brings together vector-quantized weight storage and native
MXFP4 compute, providing a practical path from compact
representations to GPU serving. Our evaluation highlights several
opportunities for MX-native inference system design.

First, the storage--accuracy trade-off is
model, task, and format-dependent which
motivates workload-specific selection of
codebook representation, MX scale granularity, and compression ratio.
Second, efficient reconstruction is workload-aware. Prefill benefits
from reconstructing FP4 operands for reuse across routed tokens,
whereas decode favors on-demand tile reconstruction. In our software
implementation, explicit reconstruction accounts for 20.4\% of the
prefill expert-layer GPU timeline, identifying indexed-FP4 operand
delivery as a target for hard\-ware--software co-design; the potential
end-to-end benefit of native support remains to be evaluated.

Third, compressed weights create deployment flexibility beyond
reducing model footprint: reclaimed HBM can support more replicas
or greater KV-cache residency. These choices must be evaluated
against service-level throughput and per-user latency objectives,
which need not improve together at high concurrency. Finally,
in our prefix-reuse workload, higher KV residency avoids redundant
recomputation of prefill and reduces P90 TTFT by up to 94.5\%. Thus, the serving
value of compression depends not only on weight bandwidth, but also
on how the saved memory is used.

Together, these findings show that
VQ's serving value extends beyond weight compression: combining
compact storage, FP4 execution, and memory-aware deployment creates
opportunities for both practical system gains and further
hardware--software co-design.

\FloatBarrier
\bibliography{references}

@article{maia26,
  title = {{Maia 200}: A Software Defined Dataflow System
             for Large-scale {AI} Acceleration},
  author = {Xu, Sherry and Heddes, Marco and others},
  journal = {arXiv preprint arXiv:2608.24664},
  year = {2026},
  doi = {10.48550/arXiv.2608.24664},
  url = {https://arxiv.org/abs/2608.24664}
}

@misc{blackwell,
  author = {{NVIDIA}},
  title = {{Blackwell SM100 GEMMs}},
  howpublished = {{NVIDIA CUTLASS Documentation}},
  url = {https://docs.nvidia.com/cutlass/latest/media/docs/cpp/blackwell_functionality.html},
  note = {Accessed: 2026-10-24}
}

@misc{mi350,
  author = {{AMD}},
  title = {{AMD Instinct MI350 Series Microarchitecture}},
  howpublished = {{AMD ROCm Documentation}},
  url = {https://rocm.docs.amd.com/en/docs-10.1.0/reference/gpu-arch/mi350.html},
  note = {Accessed: 2026-10-24}
}

@techreport{mxspec23,
  author = {{Open Compute Project}},
  title = {{OCP Microscaling Formats (MX) Specification}},
  institution = {Open Compute Project},
  number = {Version 1.0},
  month = sep,
  year = {2023},
  url = {https://www.opencompute.org/documents/ocp-microscaling-formats-mx-v1-0-spec-final-pdf}
}

@inproceedings{aqlm24,
  title = {Extreme Compression of Large Language Models
               via Additive Quantization},
  author = {Egiazarian, Vage and Panferov, Andrei
               and Kuznedelev, Denis and Frantar, Elias
               and Babenko, Artem and Alistarh, Dan},
  booktitle = {Proceedings of the 41st International
               Conference on Machine Learning},
  series = {Proceedings of Machine Learning Research},
  volume = {235},
  pages = {12284--12303},
  year = {2024},
  publisher = {PMLR},
  url = {https://proceedings.mlr.press/v235/egiazarian24a.html}
}

@inproceedings{quip24,
  title = {{QuIP\#}: Even Better {LLM} Quantization
               with {Hadamard} Incoherence and Lattice Codebooks},
  author = {Tseng, Albert and Chee, Jerry and Sun, Qingyao
               and Kuleshov, Volodymyr and De Sa, Christopher},
  booktitle = {Proceedings of the 41st International
               Conference on Machine Learning},
  series = {Proceedings of Machine Learning Research},
  volume = {235},
  pages = {48630--48656},
  year = {2024},
  publisher = {PMLR},
  url = {https://proceedings.mlr.press/v235/tseng24a.html}
}

@misc{gptvq25,
  title = {{GPTVQ}: The Blessing of Dimensionality for {LLM} Quantization},
  author = {Mart van Baalen and Andrey Kuzmin and Ivan Koryakovskiy and
            Markus Nagel and Peter Couperus and Cedric Bastoul and
            Eric Mahurin and Tijmen Blankevoort and Paul Whatmough},
  year = {2025},
  url = {https://arxiv.org/abs/2402.15319}
}

@inproceedings{vptq24,
  title = {{VPTQ}: Extreme Low-bit Vector Post-Training
               Quantization for Large Language Models},
  author = {Liu, Yifei and Wen, Jicheng and Wang, Yang
               and Ye, Shengyu and Zhang, Li Lyna
               and Cao, Ting and Li, Cheng and Yang, Mao},
  booktitle = {Proceedings of the 2024 Conference on Empirical
               Methods in Natural Language Processing},
  pages = {8181--8196},
  year = {2024},
  publisher = {Association for Computational Linguistics},
  doi = {10.18653/v1/2024.emnlp-main.467},
  url = {https://aclanthology.org/2024.emnlp-main.467/}
}

@article{mxformats23,
  author = {
    Bita Darvish Rouhani and
    Ritchie Zhao and others
  },
  title = {Microscaling Data Formats for Deep Learning},
  journal = {arXiv preprint arXiv:2310.10537},
  year = {2023},
  doi = {10.48550/arXiv.2310.10537},
  url = {https://arxiv.org/abs/2310.10537}
}

@techreport{mxspec26,
  author = {{Open Compute Project}},
  title = {{OCP MicroXcaling (MX) Specification}},
  institution = {Open Compute Project},
  number = {Version 1.1},
  month = oct,
  year = {2026},
  url = {https://www.opencompute.org/wp-content/uploads/2026/10/OCP_MicroXcaling_Spec_v1.1.0.pdf}
}

@misc{nvfp425,
  author = {
    Eduardo Alvarez and
    Omri Almog and
    Eric Chung and
    Simon Layton and
    Dusan Stosic and
    Ronny Krashinsky and
    Kyle Aubrey
  },
  title = {Introducing {NVFP4} for Efficient and Accurate
                  Low-Precision Inference},
  howpublished = {{NVIDIA} Technical Blog},
  month = jun,
  year = {2025},
  url = {https://developer.nvidia.com/blog/introducing-nvfp4-for-efficient-and-accurate-low-precision-inference/}
}

@inproceedings{finer26,
  author = {
    Andrea Fasoli and
    Monodeep Kar and
    Chi-Chun Liu and
    Swagath Venkataramani and
    Viji Srinivasan and
    Leland Chang and
    Naigang Wang
  },
  title = {Is Finer Better? {The} Limits of Microscaling Formats
               in Large Language Models},
  booktitle = {International Conference on Learning Representations
               ({ICLR})},
  year = {2026},
  doi = {10.48550/arXiv.2601.19026},
  url = {https://arxiv.org/abs/2601.19026}
}

@misc{ue5m326,
  title = {{UE5M3} {FP4} Block Scaling for Stable Language Model Pretraining},
  author = {Hu, Robert and Luschi, Carlo and Balanca, Paul},
  year = {2026},
  url = {https://arxiv.org/abs/2609.02846}
}

@misc{qwen25,
  title = {Qwen3 Technical Report},
  author = {An Yang and Anfeng Li and others},
  year = {2025},
  url = {https://arxiv.org/abs/2505.09388}
}

@article{dsr125,
  title = {DeepSeek-R1 incentivizes reasoning in LLMs through reinforcement learning},
  volume = {645},
  issn = {1476-4687},
  url = {http://dx.doi.org/10.1038/s41586-025-09422-z},
  doi = {10.1038/s41586-025-09422-z},
  number = {8081},
  journal = {Nature},
  author = {Guo, Daya and Yang, Dejian and Zhang, Haowei and others},
  year = {2025},
  month = 9,
  pages = {633–638}
}

@misc{glm526,
  author = {Z.ai},
  title = {GLM-5.2: 753B Open-Weight Mixture-of-Experts Coding Model},
  year = {2026},
  howpublished = {\url{https://huggingface.co/zai-org/GLM-5.2}}
}

@misc{lmeval23,
  author = {Gao, Leo and Tow, Jonathan and others},
  title = {A framework for few-shot language model evaluation},
  month = 12,
  year = 2023,
  url = {https://zenodo.org/records/10256836}
}
\bibliographystyle{abbrvnat}
\flushcolsend
\clearpage
\raggedcolsend
\appendix
\onecolumn
\section{Appendix}
\subsection{Scale Compression versus Coarser Scales}
\label{app:scales}
The main ablation identifies scale storage as a substantial part of the
low-bit budget. Here we examine whether spending that budget on compressed
fine-grained scales is preferable to sharing fewer, coarser scales.

Another way to reduce the scale overhead of \mbox{MXCode-PreScale} is to use coarser
scales. The stored scale granularity can be made coarser than the granularity
required by the target MX compute format. In particular, $S$ can be an integer
multiple of the native MX block size, $S=mS_{\mathrm{MX}}$ for an integer
$m\geq 1$. Each stored scale is then shared by several native MX blocks and
replicated when the weights are prepared for MX computation. This reduces the
scale cost from $b_s/S_{\mathrm{MX}}$ to $b_s/S$. The saving, however, comes
from sharing each scale across a larger group of weights. This reduces scaling
locality and can degrade reconstruction quality.

Figure~\ref{fig:sccomp} compares the two approaches at
matched compression ratios for E4M3-B16 (top) and E8M0-B32 (bottom).
For E4M3-B16, \mbox{MXCode-PreScale+} gives higher SQNR than coarser scales at every
compression ratio.
For E8M0-B32, while the advantage is smaller, at high compression ratios (e.g., 8$\times$) \mbox{MXCode-PreScale+} is still preferable over using \mbox{MXCode-PreScale} with coarser scale blocks.

Overall, compressing the scales preserves more of the pre-scaling quality
benefit than coarsening the scale granularity. It gives the better SQNR--BPE
tradeoff for E4M3-B16 and at high compression for E8M0-B32.
\vspace{10pt}

\begingroup
    \centering
    \captionsetup{type=figure,hypcap=false}
    \includegraphics[
        width=\textwidth
    ]{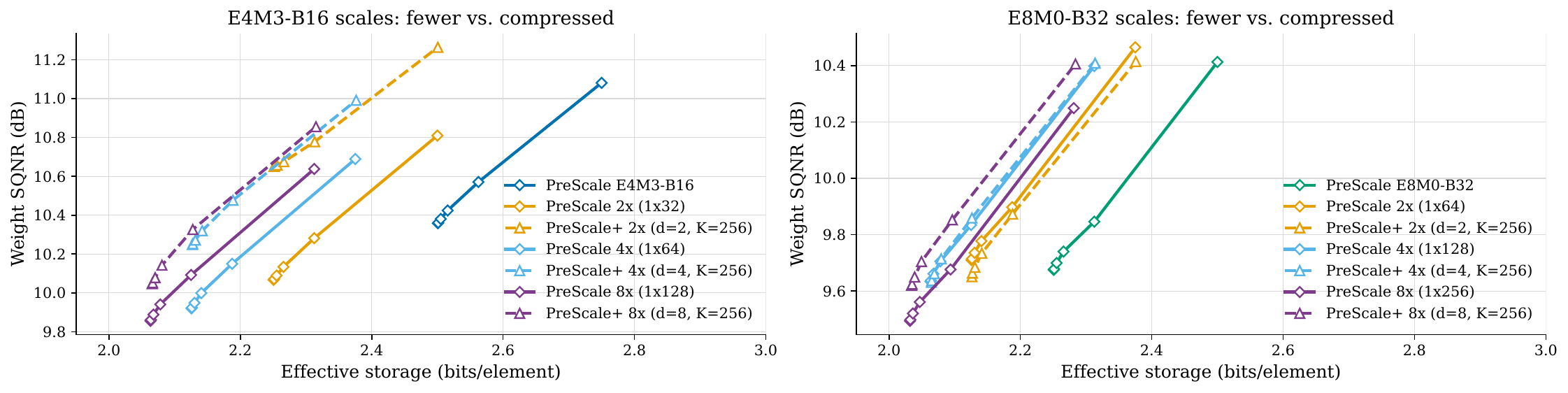}
    \caption{Scale compression versus coarser scales for E4M3-B16
    (top) and E8M0-B32 (bottom). Diamonds use coarser shared scales;
    triangles use scale VQ at matched compression ratios. Points vary
    QBlock size, with all representation metadata included in BPE.}
    \label{fig:sccomp}
\par\endgroup
\bigskip
\begingroup
\centering
\captionsetup{type=table,hypcap=false}
\caption{Weight SQNR (dB; higher is better) across Qblock sizes for 300 sampled
Qwen3-235B expert tensors. VQ uses the input-channel axis with $K=256$, $d=4$.
PreScale+ compresses scales using $K_s=256$, $d_s=8$.
E5M3 values are estimates and E4M3 values are proxies.
$\Delta=\mathrm{SQNR}_{\mathrm{E5M3}}-\mathrm{SQNR}_{\mathrm{E4M3}}$,
computed from displayed values; positive values favor E5M3.}
\label{tab:sqnr}
\begingroup
\fontsize{8}{9}\selectfont
\setlength{\tabcolsep}{3pt}
\renewcommand{\arraystretch}{1.10}
\begin{tabular*}{\textwidth}{@{\extracolsep{\fill}}cl*{9}{c}@{}}
\toprule
 & \textbf{Method} & \multicolumn{3}{c}{\textbf{MXCode PostScale}} & \multicolumn{3}{c}{\textbf{MXCode PreScale}} & \multicolumn{3}{c}{\textbf{MXCode PreScale+}} \\
\cmidrule(lr){3-5}\cmidrule(lr){6-8}\cmidrule(lr){9-11}
 & \textbf{Compute format} & \textbf{E5M3} & \textbf{E4M3} & \textbf{$\Delta$} & \textbf{E5M3} & \textbf{E4M3} & \textbf{$\Delta$} & \textbf{E5M3} & \textbf{E4M3} & \textbf{$\Delta$} \\
\midrule
 & \textbf{Qblock size} & \multicolumn{9}{c}{\textbf{Weight SQNR (dB)}} \\
\midrule
\multirow{8}{*}{\rotatebox[origin=c]{90}{\textbf{Qwen3 (235B)}}} & Full tensor & 7.035 & 7.046 & $-0.011$ & 10.349 & 10.359 & $-0.010$ & 10.049 & 10.047 & $+0.002$ \\
 & $2048\times2048$ & 7.795 & 7.798 & $-0.003$ & 10.363 & 10.361 & $+0.002$ & 10.051 & 10.052 & $-0.001$ \\
 & $1024\times1024$ & 8.795 & 8.784 & $+0.011$ & 10.384 & 10.380 & $+0.004$ & 10.079 & 10.078 & $+0.001$ \\
 & $512\times512$ & 9.342 & 9.317 & $+0.025$ & 10.429 & 10.424 & $+0.005$ & 10.145 & 10.142 & $+0.003$ \\
 & $256\times256$ & 9.802 & 9.796 & $+0.006$ & 10.572 & 10.571 & $+0.001$ & 10.327 & 10.326 & $+0.001$ \\
 & $128\times128$ & 10.447 & 10.445 & $+0.002$ & 11.085 & 11.081 & $+0.004$ & 10.856 & 10.855 & $+0.001$ \\
 & $64\times128$ & 11.092 & 11.091 & $+0.001$ & 11.632 & 11.632 & $0.000$ & 11.415 & 11.411 & $+0.004$ \\
 & $64\times64$ & 12.125 & 12.125 & $0.000$ & 12.668 & 12.665 & $+0.003$ & 12.450 & 12.447 & $+0.003$ \\
\bottomrule
\end{tabular*}
\endgroup
\par\endgroup
\clearpage
\subsection{Evaluation of E5M3-B16-FP4}
\label{app:e5m3}
We additionally evaluate E5M3-B16-FP4 as an alternative scale format
for MXCode. Its wider block-scale dynamic range eliminates the need
for an additional tensor-level scale while retaining 16-element
scaling granularity. The proxy-based SQNR comparison in
Table~\ref{tab:sqnr} suggests similar weight
reconstruction quality across formats, with estimated differences
of at most 0.025\,dB across the evaluated methods and Qblock sizes.
The end-to-end W4A4 results in
Table~\ref{tab:e5m3} likewise show
comparable accuracy on Qwen3 (235B). E5M3 PreScale and PreScale+
achieve average accuracies of 88.46\% and 88.20\% at 2.56 and
2.13 BPE, respectively, compared with 88.93\% for native FP4
at 4.50 BPE. Across the three MXCode schemes, E5M3 average accuracy
exceeds the corresponding E4M3 result by 0.14--0.37 percentage
points, although individual benchmark results vary. Together,
these results support E5M3 as an alternative that preserves
MXCode's storage--quality trade-offs without requiring a
tensor-level scale.

\begingroup
\centering
\captionsetup{type=table,hypcap=false}
\caption{W4A4 evaluation on Qwen3 (235B), comparing E5M3-B16-FP4
and E4M3-B16-FP4. Accuracy and standard errors are percentages.
$\Delta$ denotes E5M3 minus E4M3 accuracy in percentage points;
positive values favor E5M3. Weight storage includes indices,
codebooks, and scales, as applicable. Average accuracy weights
the ten benchmarks equally.}
\label{tab:e5m3}
\begingroup
\fontsize{8}{9}\selectfont
\setlength{\tabcolsep}{2pt}
\renewcommand{\arraystretch}{1.10}
\begin{tabular*}{\textwidth}{@{\extracolsep{\fill}}l*{12}{c}@{}}
\toprule
\textbf{Method} & \multicolumn{3}{c}{\textbf{Native FP4}} & \multicolumn{3}{c}{\textbf{MXCode PostScale}} & \multicolumn{3}{c}{\textbf{MXCode PreScale}} & \multicolumn{3}{c}{\textbf{MXCode PreScale+}} \\
\cmidrule(lr){2-4}\cmidrule(lr){5-7}\cmidrule(lr){8-10}\cmidrule(lr){11-13}
\textbf{Compute format} & \textbf{E5M3} & \textbf{E4M3} & \textbf{$\Delta$} & \textbf{E5M3} & \textbf{E4M3} & \textbf{$\Delta$} & \textbf{E5M3} & \textbf{E4M3} & \textbf{$\Delta$} & \textbf{E5M3} & \textbf{E4M3} & \textbf{$\Delta$} \\
\midrule
\textbf{Weight (BPE)} & 4.50 & 4.50 & -- & 2.06 & 2.06 & -- & 2.56 & 2.56 & -- & 2.13 & 2.13 & -- \\
\midrule
MMLU-Pro & \shortstack{82.45\\\textcolor{gray}{\(\pm 0.34\)}} & \shortstack{82.41\\\textcolor{gray}{\(\pm 0.34\)}} & $+0.04$ & \shortstack{79.04\\\textcolor{gray}{\(\pm 0.36\)}} & \shortstack{79.18\\\textcolor{gray}{\(\pm 0.36\)}} & $-0.14$ & \shortstack{79.75\\\textcolor{gray}{\(\pm 0.35\)}} & \shortstack{80.34\\\textcolor{gray}{\(\pm 0.35\)}} & $-0.59$ & \shortstack{80.54\\\textcolor{gray}{\(\pm 0.35\)}} & \shortstack{80.54\\\textcolor{gray}{\(\pm 0.35\)}} & $0.00$ \\
RULER & \shortstack{95.11\\\textcolor{gray}{\(\pm 0.13\)}} & \shortstack{95.05\\\textcolor{gray}{\(\pm 0.14\)}} & $+0.06$ & \shortstack{94.31\\\textcolor{gray}{\(\pm 0.14\)}} & \shortstack{94.16\\\textcolor{gray}{\(\pm 0.15\)}} & $+0.15$ & \shortstack{94.89\\\textcolor{gray}{\(\pm 0.14\)}} & \shortstack{95.01\\\textcolor{gray}{\(\pm 0.14\)}} & $-0.12$ & \shortstack{94.97\\\textcolor{gray}{\(\pm 0.14\)}} & \shortstack{95.00\\\textcolor{gray}{\(\pm 0.14\)}} & $-0.03$ \\
DROP & \shortstack{88.49\\\textcolor{gray}{\(\pm 0.33\)}} & \shortstack{88.56\\\textcolor{gray}{\(\pm 0.33\)}} & $-0.07$ & \shortstack{86.69\\\textcolor{gray}{\(\pm 0.35\)}} & \shortstack{86.98\\\textcolor{gray}{\(\pm 0.34\)}} & $-0.29$ & \shortstack{87.71\\\textcolor{gray}{\(\pm 0.34\)}} & \shortstack{86.54\\\textcolor{gray}{\(\pm 0.35\)}} & $+1.17$ & \shortstack{87.56\\\textcolor{gray}{\(\pm 0.34\)}} & \shortstack{87.69\\\textcolor{gray}{\(\pm 0.34\)}} & $-0.13$ \\
GSM8K & \shortstack{97.35\\\textcolor{gray}{\(\pm 0.44\)}} & \shortstack{96.89\\\textcolor{gray}{\(\pm 0.48\)}} & $+0.46$ & \shortstack{96.59\\\textcolor{gray}{\(\pm 0.50\)}} & \shortstack{96.66\\\textcolor{gray}{\(\pm 0.49\)}} & $-0.07$ & \shortstack{97.12\\\textcolor{gray}{\(\pm 0.46\)}} & \shortstack{97.12\\\textcolor{gray}{\(\pm 0.46\)}} & $0.00$ & \shortstack{96.97\\\textcolor{gray}{\(\pm 0.47\)}} & \shortstack{96.36\\\textcolor{gray}{\(\pm 0.52\)}} & $+0.61$ \\
IFEval & \shortstack{83.55\\\textcolor{gray}{\(\pm 1.60\)}} & \shortstack{84.47\\\textcolor{gray}{\(\pm 1.56\)}} & $-0.92$ & \shortstack{77.45\\\textcolor{gray}{\(\pm 1.80\)}} & \shortstack{79.48\\\textcolor{gray}{\(\pm 1.74\)}} & $-2.03$ & \shortstack{81.89\\\textcolor{gray}{\(\pm 1.66\)}} & \shortstack{81.70\\\textcolor{gray}{\(\pm 1.66\)}} & $+0.19$ & \shortstack{81.33\\\textcolor{gray}{\(\pm 1.68\)}} & \shortstack{81.33\\\textcolor{gray}{\(\pm 1.68\)}} & $0.00$ \\
HumanEval & \shortstack{96.95\\\textcolor{gray}{\(\pm 1.35\)}} & \shortstack{96.95\\\textcolor{gray}{\(\pm 1.35\)}} & $0.00$ & \shortstack{94.51\\\textcolor{gray}{\(\pm 1.78\)}} & \shortstack{90.24\\\textcolor{gray}{\(\pm 2.32\)}} & $+4.27$ & \shortstack{97.56\\\textcolor{gray}{\(\pm 1.21\)}} & \shortstack{98.17\\\textcolor{gray}{\(\pm 1.05\)}} & $-0.61$ & \shortstack{98.17\\\textcolor{gray}{\(\pm 1.05\)}} & \shortstack{95.73\\\textcolor{gray}{\(\pm 1.58\)}} & $+2.44$ \\
MATH-500 & \shortstack{96.20\\\textcolor{gray}{\(\pm 0.86\)}} & \shortstack{97.00\\\textcolor{gray}{\(\pm 0.76\)}} & $-0.80$ & \shortstack{94.80\\\textcolor{gray}{\(\pm 0.99\)}} & \shortstack{95.20\\\textcolor{gray}{\(\pm 0.96\)}} & $-0.40$ & \shortstack{95.80\\\textcolor{gray}{\(\pm 0.90\)}} & \shortstack{95.60\\\textcolor{gray}{\(\pm 0.92\)}} & $+0.20$ & \shortstack{95.80\\\textcolor{gray}{\(\pm 0.90\)}} & \shortstack{96.00\\\textcolor{gray}{\(\pm 0.88\)}} & $-0.20$ \\
GPQA-Diamond & \shortstack{69.19\\\textcolor{gray}{\(\pm 3.29\)}} & \shortstack{67.17\\\textcolor{gray}{\(\pm 3.35\)}} & $+2.02$ & \shortstack{66.16\\\textcolor{gray}{\(\pm 3.37\)}} & \shortstack{64.65\\\textcolor{gray}{\(\pm 3.41\)}} & $+1.51$ & \shortstack{67.17\\\textcolor{gray}{\(\pm 3.35\)}} & \shortstack{64.14\\\textcolor{gray}{\(\pm 3.42\)}} & $+3.03$ & \shortstack{64.65\\\textcolor{gray}{\(\pm 3.41\)}} & \shortstack{64.14\\\textcolor{gray}{\(\pm 3.42\)}} & $+0.51$ \\
MGSM & \shortstack{88.25\\\textcolor{gray}{\(\pm 0.60\)}} & \shortstack{86.51\\\textcolor{gray}{\(\pm 0.63\)}} & $+1.74$ & \shortstack{87.05\\\textcolor{gray}{\(\pm 0.61\)}} & \shortstack{90.69\\\textcolor{gray}{\(\pm 0.55\)}} & $-3.64$ & \shortstack{89.67\\\textcolor{gray}{\(\pm 0.57\)}} & \shortstack{91.16\\\textcolor{gray}{\(\pm 0.54\)}} & $-1.49$ & \shortstack{89.60\\\textcolor{gray}{\(\pm 0.57\)}} & \shortstack{90.15\\\textcolor{gray}{\(\pm 0.56\)}} & $-0.55$ \\
MBPP & \shortstack{91.80\\\textcolor{gray}{\(\pm 1.23\)}} & \shortstack{94.40\\\textcolor{gray}{\(\pm 1.03\)}} & $-2.60$ & \shortstack{91.40\\\textcolor{gray}{\(\pm 1.26\)}} & \shortstack{89.40\\\textcolor{gray}{\(\pm 1.38\)}} & $+2.00$ & \shortstack{93.00\\\textcolor{gray}{\(\pm 1.14\)}} & \shortstack{91.80\\\textcolor{gray}{\(\pm 1.23\)}} & $+1.20$ & \shortstack{92.40\\\textcolor{gray}{\(\pm 1.19\)}} & \shortstack{91.40\\\textcolor{gray}{\(\pm 1.26\)}} & $+1.00$ \\
\midrule
\textbf{Average accuracy} & 88.93 & 88.94 & $-0.01$ & 86.80 & 86.66 & $+0.14$ & 88.46 & 88.16 & $+0.30$ & 88.20 & 87.83 & $+0.37$ \\
\bottomrule
\end{tabular*}
\endgroup
\bigskip
\par\endgroup

\end{document}